\documentclass[
dvipsnames, table,   
format=sigconf,
anonymous=false,     
review=false,        
authorversion=false, 
screen=true,         
nonacm=true,         
]{acmart}

\usepackage{custom_commands}
\usepackage{custom_aliases}
\usepackage{graphics}
\usepackage{bm}
\usepackage{breqn}
\usepackage{setspace}
\usepackage[thinc]{esdiff}

\newcommand{\mathsym}[1]{{}}
\newcommand{\unicode}[1]{{}}

\newcommand{\Rea}{\mathbb{R}}

\copyrightyear{2022}
\acmYear{2022}
\setcopyright{acmlicensed}\acmConference[GECCO '22 Companion]{Genetic and Evolutionary Computation Conference Companion}{July 9--13, 2022}{Boston, MA, USA}
\acmBooktitle{Genetic and Evolutionary Computation Conference Companion (GECCO '22 Companion), July 9--13, 2022, Boston, MA, USA}
\acmPrice{15.00}
\acmDOI{10.1145/3520304.3533972}
\acmISBN{978-1-4503-9268-6/22/07}

\begin{document}


\title{Mono-surrogate vs Multi-surrogate in Multi-objective Bayesian Optimisation}


\author{Tinkle Chugh}
\email{t.chugh@exeter.ac.uk}
\orcid{0000-0001-5123-8148}
\affiliation{%
  \department{Department of Computer Science}
  \institution{University of Exeter}
  \city{Exeter}
  \country{United Kingdom}
}

\renewcommand{\shortauthors}{Chugh}
\begin{abstract}

Bayesian optimisation (BO) has been widely used to solve problems with expensive function evaluations. In multi-objective optimisation problems, BO aims to find a set of approximated Pareto optimal solutions. There are typically two ways to build surrogates in multi-objective BO: One surrogate by aggregating objective functions (by using a scalarising function, also called mono-surrogate approach) and multiple surrogates (for each objective function, also called multi-surrogate approach). In both approaches, an acquisition function (AF) is used to guide the search process. Mono-surrogate has the advantage that only one model is used, however, the approach has two major limitations. Firstly, the fitness landscape of the scalarising function and the objective functions may not be similar. Secondly, the approach assumes that the scalarising function distribution is Gaussian, and thus a closed-form expression of the AF can be used. In this work, we overcome these limitations by building a surrogate model for each objective function and show that the scalarising function distribution is not Gaussian. We approximate the distribution using Generalised extreme value distribution. The results and comparison with existing approaches on standard benchmark and real-world optimisation problems show the potential of the multi-surrogate approach. 

\end{abstract}

%
%



\maketitle


\section{Introduction}
Many real-world optimisation problems involve multiple conflicting objectives to be achieved. These problems are called multi-objective optimisation problems (MOPs). There is no single solution to such problems because of the conflicting nature between the objectives. 
The solutions to such problems are known as Pareto optimal solutions, and represent the trade-off between objectives \cite{Miettinen1999}. We define a multi-objective optimisation problem (MOP) as:
\begin{equation*}\label{MOOP}
  \text{minimise} \enspace \bff =  \left(f_1(\bx),\dotsc,f_m(\bx)\right)\qquad\mbox{subject to} \quad \bx \in S 
\end{equation*}
%
with $m \geq 2$ objective functions $f_i(\bx)\colon S\to\Rea$. 
The (nonempty) feasible space $S$ is a subset of the decision space $\Rea^n$ and consists of decision vectors $\bx=(x_1,\ldots,x_n)^T$. In solving such problems, usually, the aim is to find an approximated set of Pareto optimal solutions.  

In many real-world MOPs, the objective functions rely on computationally expensive evaluations. Such problems are usually black-box optimisation problems without any closed-form for the objective functions. Bayesian optimisation (BO) can be used to alleviate the computational cost and to find an approximated set of Pareto optimal solutions in the least number of function evaluations. These methods rely on a Bayesian model as the surrogate (or metamodel) of the objective functions and find promising decision vectors by optimising an acquisition function. The Bayesian model is usually a Gaussian process because it provides a meaningful quantification of uncertainty, which is then used in optimising the acquisition function. The acquisition function balances both exploration and exploitation in guiding the search process. A single objective BO with expected improvement (EI) as the acquisition function was first used in \cite{Mockus1975}. 
Emmerich et al.\ \cite{Emmerich2004,Emmerich2005a,Emmerich2006,Emmerich2011,Yang2019} extended the EI to expected hypervolume improvement (EHVI) in multi-objective Bayesian optimisation. Recently, a closed-form expression of EHVI was proposed for batch multi-objective Bayesian optimisation \cite{NEURIPS2020_6fec24ea}.

In multi-objective BO, there are typically two different approaches to build a Bayesian model. In the first one, the models are built for each objective function and an acquisition function utilising these models is then used to find promising decision vectors. This approach is called multi-surrogate approach. The multi-objective BO with EHVI is a multi-surrogate approach. In the second one, a single Bayesian model is built after aggregating the objective functions. This approach is called mono-surrogate approach. The well-known ParEGO \cite{Knowles2006} algorithm comes under the second category. The second approach reduces the number of objectives from $m$ to one. Moreover, a single objective acquisition function can be used in the mono-surrogate approach. The computational complexity of the first approach is at most $O(mN^3)$ and of the second approach is at most $O(N^3)$, where $N$ is the size of the data set. 

The ParEGO converts the multiple objectives into a single objective by utilising the augmented weighted Tchebycheff (TCH) \cite{Wierzbicki1980} as the scalarising function. A Gaussian process model is then built on the scalarising function, which is then used in optimising the EI to find the next promising decision vector. Although the mono-surrogate approach is appealing, it has two major limitations. The first one is that the fitness landscape of the scalarising function and the objective functions may not be similar. In other words, a promising decision vector by using the surrogate on the scalarising function may not be promising for the underlying objective functions. The second limitation is that the approach assumes that the resulting scalarising function is Gaussian and thus a closed-form expression of the EI can be used. 

We show that the distribution of the scalarising function after building individual models for each objective function is not Gaussian. 
Therefore, we approximate the distribution of the scalarising function using the Generalised extreme value (GEV) distribution \cite{GEV,Haan2006}. Particularly, we use the type I family of distributions in the generalised extreme value theory to approximate the distribution. The GEV distribution is then used in optimising the expected improvement to find the next promising decision vector.

We compare mono-surrogate EI (as in ParEGO), multi-surrogate EI (this work) and multi-surrogate EHVI (as in EHVI-EGO \cite{Yang2019}) approaches on standard benchmark and real-world optimisation problems with a different number of objectives and decision variables. The results on benchmark and real-world multi-objective optimisation problems clearly show the potential of the proposed work. To be summarised, the contributions of this work are as follows:
\begin{enumerate}
    \item We show that the weighted Tchebycheff function (TCH) as the scalarising function after building independent models for each objective function is not Gaussian.
    \item We approximate the distribution of the TCH with the type I family of the Generalised extreme value theory.
    \item We solve a real-world multi-objective free-radical polymerisation optimisation problem by using the multi-surrogate EI approach.
\end{enumerate}

The rest of the paper is structured as follows. In Section 2, we provide a background to the Bayesian optimisation. In Section 3, we provide the main details of the proposed approach by comparing it with a mono-surrogate approach. In Section 4, we conduct numerical experiments and show the results on benchmark and real-world multi-objective optimisation problems. Finally, we conclude and mention the future research direction in Section 5. 

\section{Bayesian optimisation}


In multi-objective BO, the input is the data set $\{ (\bx_i, \bff(\bx_i))\}_{i=1}^N$ of size $N$. This data set can be obtained with a design of experiment technique e.g.\ Latin Hypercube sampling \cite{Mckay2000}. As mentioned in the introduction, there are typically two approaches to build surrogate models on the data set: A single Bayesian model (mono-surrogate approach) or multiple Bayesian models (multi-surrogate approach). The Gaussian process models are the most commonly used Bayesian models in BO. They are non-parametric and provide uncertainties in predictions, which makes them different from other non-Bayesian and parametric models. The uncertainty is then used in the acquisition function in finding the promising decision vector. A Gaussian process ($\mGP$) can be defined with a multivariate normal distribution \cite{Rasmussen2006}:
\begin{align*}
  \mathbf{f} \sim \mathcal{N}(\bm{\mu}, K),
  \label{eq:Kriging}
\end{align*}
where $\bm{\mu}$ is the mean vector and $K$ is the covariance matrix. Without loss of generality, we use a prior of zero mean. A covariance function (or kernel) is used to get the covariance matrix. In this work, we used a Gaussian (or RBF or squared exponential) kernel with automatic relevant determination \cite{Chugh_2019_LOD,Palar_AIAA}:
\begin{align*}
    \kappa(\mathbf{x}, \mathbf{x'}, \mathbf{\Theta}) = \sigma_f^2
    \exp\left(-\frac{1}{2}\sum\limits_{j=1}^n\frac{|x_{j} - x_{j}'|^2}{l_j^2}\right) + \sigma_n^2 \delta_{\mathbf{x}\mathbf{x'}},
\end{align*} 
where $\Theta = (\sigma_f, l_1,\ldots,l_n,\sigma_n)$ is the set of hyperparameters and $\delta_{\mathbf{x}\mathbf{x'}}$ is the Kronecker delta function. The notation $|x_{j} - x_{j}'|$ represents the Euclidean distance between $x_j$ and $x_j'$. The hyperparameters can be estimated by maximising the marginal likelihood function \cite{Rasmussen2006}:
\begin{align*}
    p(\mathbf{f}|X,\mathbf{\Theta}) = \frac{1}{|2\pi K |^{\frac{1}{2}}} \exp \left\{
  -\tfrac{1}{2} \mathbf{f}^T K^{-1} \mathbf{f}
  \right\}.
\end{align*}
The posterior predictive distribution at new point $\bx^*$after training the model is also Gaussian:
\begin{align*}
\begin{split}
 p\left(f^* | \mathbf{x^*},X,\mathbf{f},\mathbf{\Theta} \right) =& 
 \mathcal{N}\Big( \bm{\kappa}(\mathbf{x^*}, X) K ^{-1} \mathbf{f},\\
 &\kappa(\mathbf{x^*}, \mathbf{x^*}) - \bm{\kappa}(\mathbf{x^*}, X) ^T K ^{-1} \bm{\kappa}(X, \mathbf{x^*}) \Big).
 \end{split}
\end{align*}



%

The acquisition function determines the next decision vector to be evaluated. The expected improvement (EI) is one of the well-known and widely used acquisition functions. It measures the amount of improvement over the current best objective value and balances both exploitation and exploration. For a minimisation problem, the improvement over the best evaluated function value $f'(\bx)$ is:
\begin{align*}
    I(\bx) = \max (0, f'(\bx) - f).
\end{align*}
The expected improvement can then be estimated as:
\begin{align*}
    \alpha_{EI}(\bx) = \int_{-\infty}^{f'(\bx)}I(x) df
\end{align*}
As the posterior is Gaussian, the expected improvement has a closed-form expression:
\begin{align*}
    \alpha_{EI}(\bx) = (f'(\bx) - \mu(\bx))\Phi\Big(\frac{f'(\bx) - \mu(\bx)}{\sigma(\bx)}\Big) + \sigma (\bx) \phi\Big(\frac{f'(\bx) - \mu(\bx)}{\sigma(\bx)}\Big),
\end{align*}
where $\mu(x)$ and $\sigma(x)$ is the posterior mean and standard deviation, respectively and $\Phi(\cdot)$ and $\phi(\cdot)$ are cumulative and probability distribution function of standard normal distribution, respectively.

In mono-surrogate multi-objective BO, this formulation of EI can be used as the objective functions are aggregated into a single function. In the multi-surrogate approach, EHVI, which is an extension of EI for multiple objectives can be used. In this work, we focus on EI after using multiple surrogates. The EI is usually a multimodal function and therefore, a suitable optimiser e.g.\ an evolutionary algorithm is often used to optimise it to find the next decision vector, which is then evaluated with expensive objective function and added to the data set. This process continues until a termination criterion is met. Algorithm \ref{alg:BO} outlines these steps.



\begin{algorithm} [t!]
\caption{Bayesian optimisation}
\label{alg:BO}
\begin{algorithmic}[]
    \State \textbf{Input}: 
        Data Set $D = \{ (\bx_i,\bff_i) \}_{i=1}^N$
    \State \textbf{Output}: 
        Evaluated solutions
\end{algorithmic}
\begin{algorithmic}[1]
    \While{Termination criterion is not met}
    \State Train the $\mGP$ models on the data set
    \State Optimise the acquisition function i.e.\ $\bx^* \leftarrow \argmax_{\bx} EI(x)$ 
    \State Evaluate $\bx^*$ and add to the data set
    \EndWhile
\end{algorithmic}
\end{algorithm}

\section{Mono-surrogate vs Multi-surrogate}
The mono-surrogate approach is one of the well-known approaches in Bayesian optimisation. We start by introducing the weighted Tchebycheff (TCH) function \cite{Wierzbicki1980, Miettinen1999, mbore_george} in a widely used mono-surrogate approach (the ParEGO algorithm). The TCH function is defined as:
\begin{align*}
    g = \max_{i} \big( w_i (f_i - z_i)\big),
\end{align*}
where $\bww$ is the weight vector and $\mathbf{z}$ is the ideal objective vector (or minimum of objective function values). In ParEGO, an augmented Tchebycheff formulation was used to obtain properly Pareto optimal solutions \cite{Miettinen1999}. If the objective function values at the current iteration are normalised between 0 and 1, then $\mathbf{z}$ is a vector of zeros. Given a data set with decision variable and objective function values, the TCH converts the multiple objective function values into a single value. A Gaussian process model is then built on the resulting data set, which is then used in optimising the expected improvement. Recently, three other scalarising functions called hypervolume improvement, dominance ranking and sign distance were proposed in \cite{Alma2017} and used in the framework of the mono-surrogate approach.  

In this work, we use a multi-surrogate approach by building independent models on each objective function. These independent models are then used to build a probabilistic model for the TCH function, which is then used in optimising the EI. We start by providing a simple example to compare both approaches with TCH as the scalarising function.

\begin{figure*}
    \centering
    \includegraphics[width=0.33\textwidth]{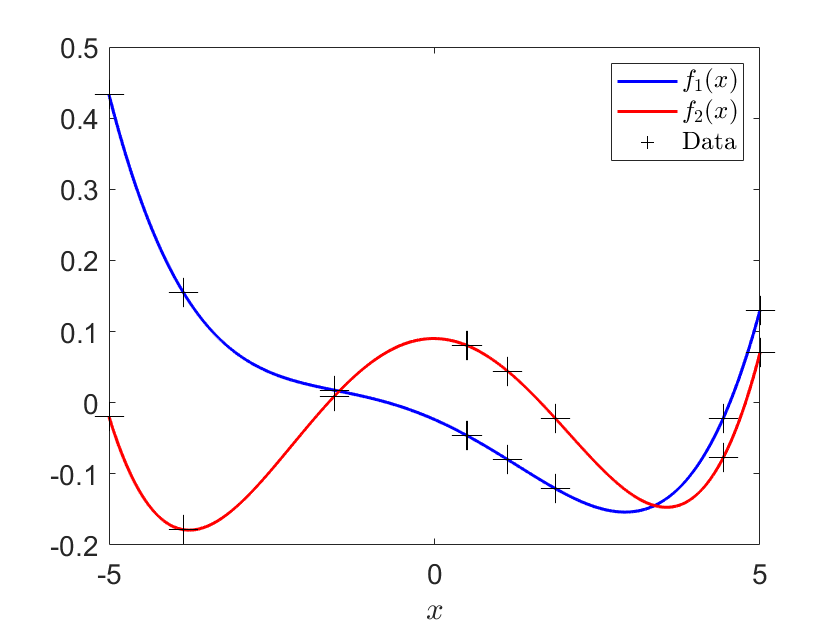}
    \includegraphics[width=0.33\textwidth]{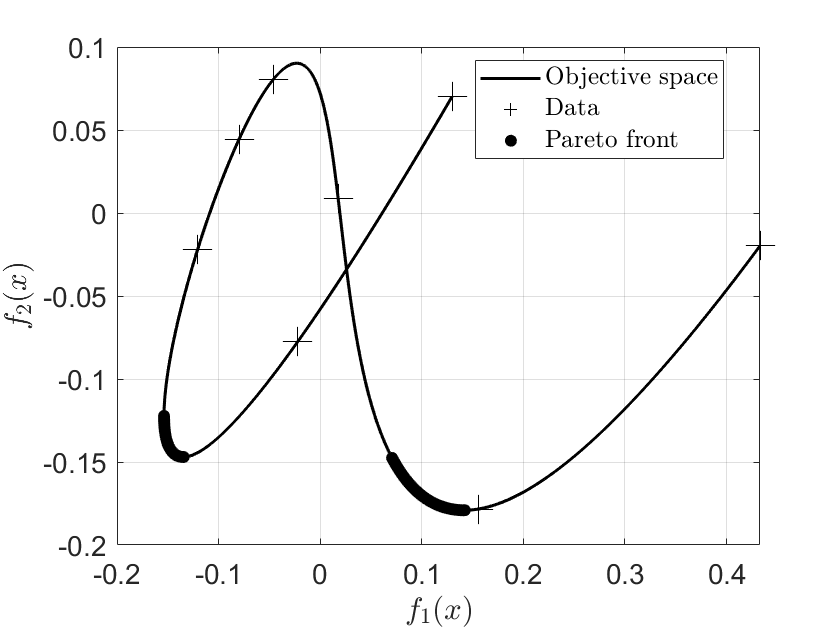}
    \includegraphics[width=0.33\textwidth]{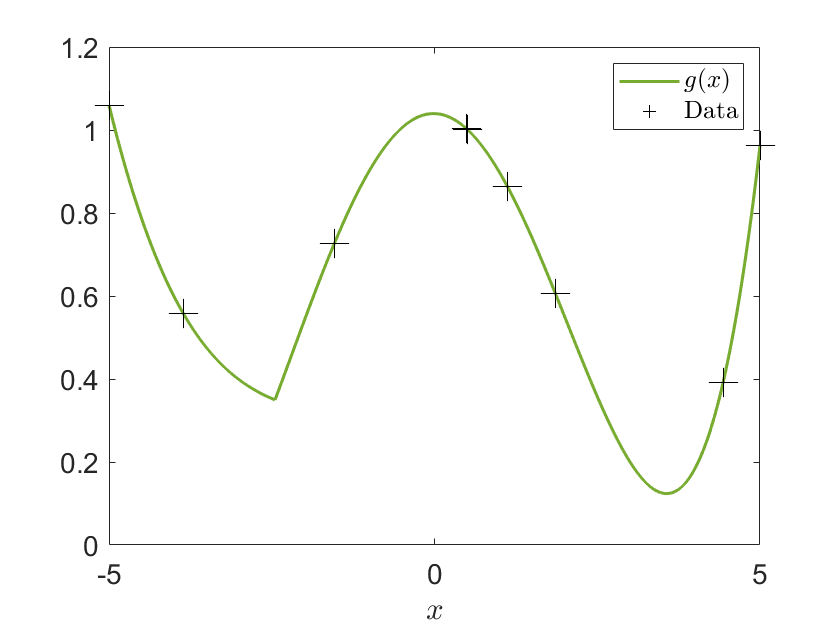}
    \caption{A bi-objective optimisation problem. Both objectives are to be minimised. The data is shown in '+' in the left figure. The Pareto front and the objective space are shown in the middle figure. The resulting scalarising function with the data set is shown in the right figure.}
    \label{fig:demo_1}
\end{figure*}
\begin{figure*}
    \centering
    \includegraphics[width=0.45\textwidth]{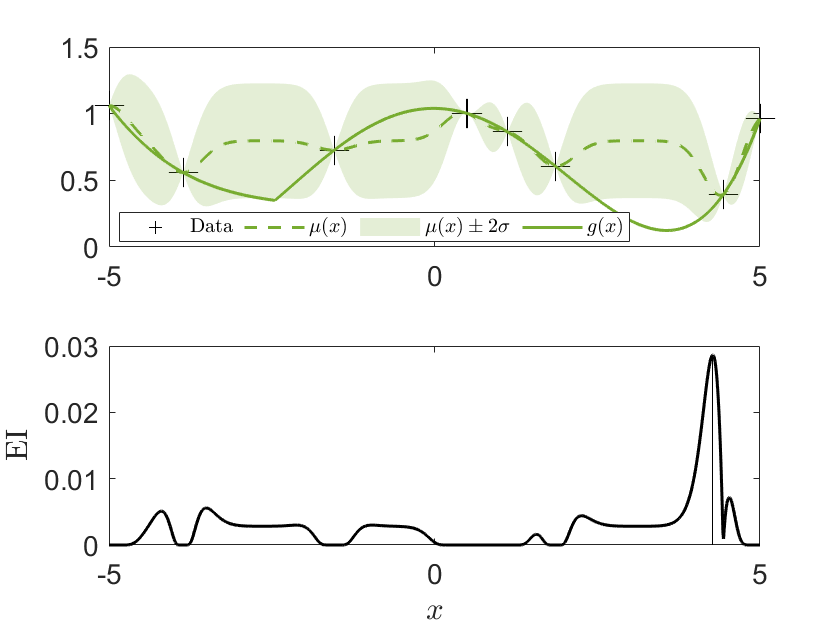}
    \includegraphics[width=0.45\textwidth]{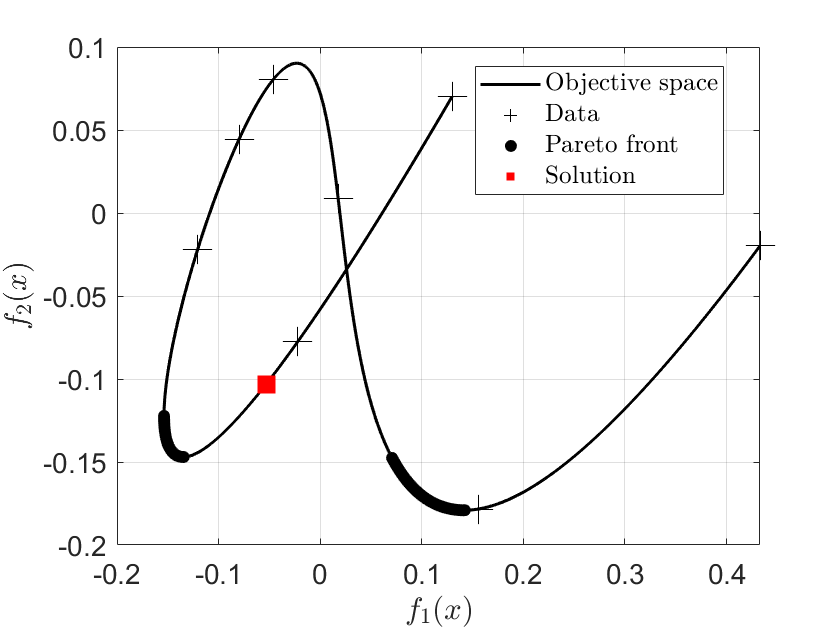}
    \caption{A Gaussian process model and the landscape of EI on the weighted Tchebycheff function (left figure). The evaluated solution (shown as the square marker) in the objective space after maximising the EI (right figure).}
    \label{fig:demo_2}
\end{figure*}

\begin{figure*}
    \centering
    \includegraphics[width=0.45\textwidth]{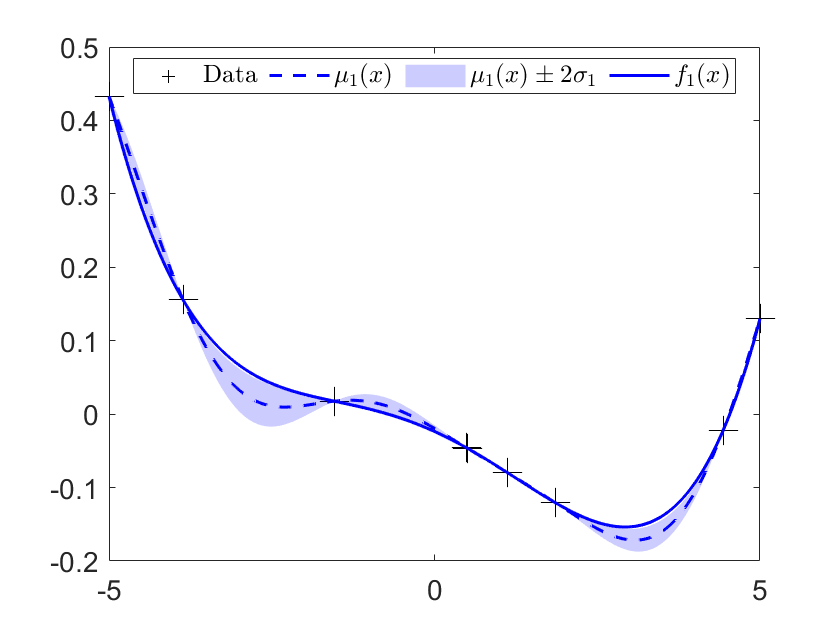}
    \includegraphics[width=0.45\textwidth]{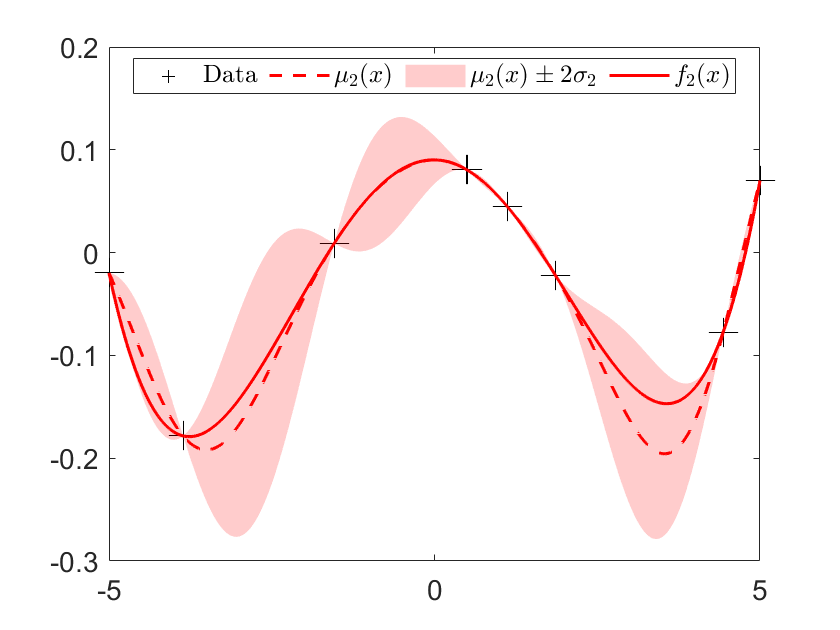}
    \caption{A Gaussian process model for each objective function.}
    \label{fig:demo_3}
\end{figure*}

\begin{figure*}
    \centering
    \includegraphics[width=0.4\textwidth]{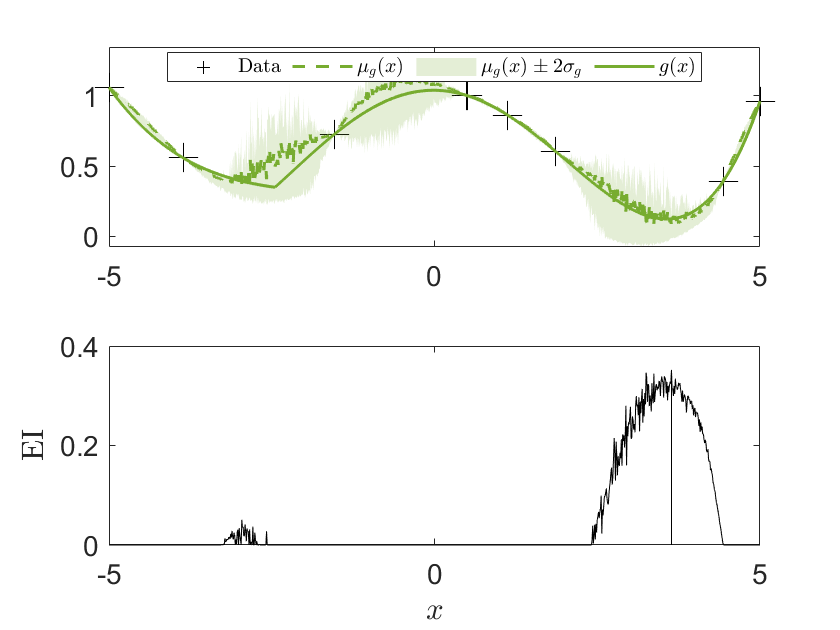}
    \includegraphics[width=0.4\textwidth]{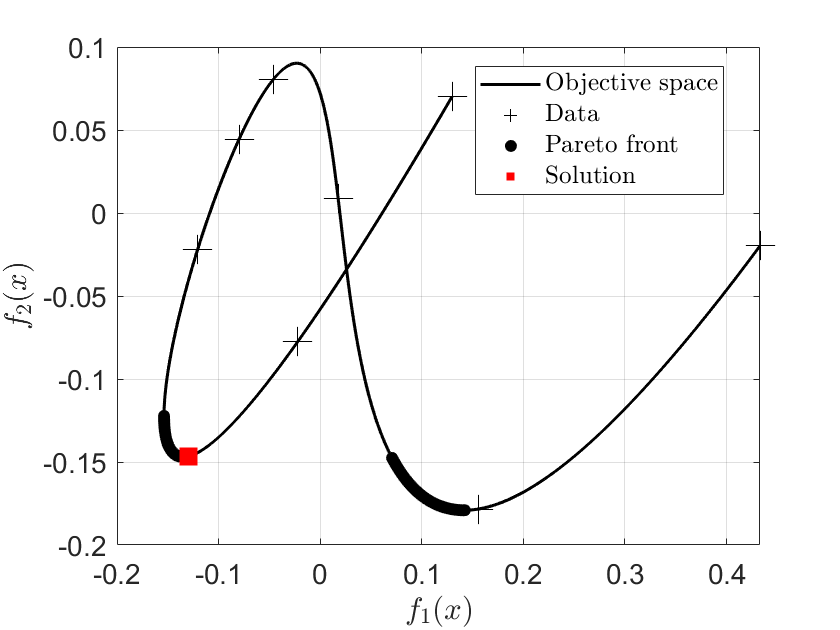}
    \caption{A Gaussian process model and the landscape of EI on the weighted Tchebycheff function after building independent models for each objective function (left figure). The evaluated solution (shown as the square marker) in the objective space after maximising the EI (right figure).}
    \label{fig:demo_4}
\end{figure*}

Consider two objective functions $f_1$ and $f_2$ shown in Figure \ref{fig:demo_1}. We assume that these functions are black box and their analytical forms are not available. However, we have some data set (with decision variable and objective functions values) shown as $+$ in the figure. The true or underlying TCH function is also shown in the figure. As mentioned, there are two ways to build surrogate models: mono-surrogate and multi-surrogate. In the mono-surrogate approach, we get the TCH values after aggregating two objective values (right plot in Figure \ref{fig:demo_1}). We then build a Gaussian process model on it as shown in Figure \ref{fig:demo_2}. We then maximise the EI to get the next decision variable value i.e.\ $x^* = \argmax \text{EI}$. The location of $x^*$ can be seen in Figure \ref{fig:demo_2}. We then evaluate the $x^*$ with the objective functions. The resulting objective function values are shown as square in the right in Figure \ref{fig:demo_2}. In this way, the mono-surrogate approach tries to find a potential decision vector for the scalarising function and not for the underlying objective functions. To address this concern, we can build independent models for objective functions.  


The predictions with uncertainty estimates after building independent models are shown in Figure \ref{fig:demo_3}. The challenge is how to use these models to find the distribution of the TCH as the scalarising function. We start to address this challenge by showing that the resulting scalarising function after building independent models on objective functions is not Gaussian. The weighted Tchebycheff function is defined as:
\begin{equation*}
    g = \max_i\big( w_i(f_i - z_i) \big),
\end{equation*}
where $f_i$ is Gaussian i.e.\ $f_i \sim \mathcal{N}(\mu_i, \sigma_i^2)$; $\mu_i$ and $\sigma_i$ are the posterior predictive means and standard deviations, respectively. The above formulation can be written as \cite{Lemons2002,Mood1974}:
\begin{align}
     g \sim& \max_i \mathcal{N}\Big( w_i(\mu_i - z_i), w_i^2 \sigma_i^2\Big)
     \label{eq:g_pdf1}
\end{align}
After some rearrangements, the distribution shown above can be written in the following closed form expression \cite{Azzalini1985,Nadarajah2008,Atanu_TEVC}: 

\begin{dmath}
    p(g) =\sum_{i=1}^m \frac{1}{w_i\sigma_i} \times \frac{\phi\Big( \frac{g - w_i(\mu_i - z_i)}{w_i\sigma_i} \Big)}{\Phi\Big( \frac{g - w_i(\mu_i - z_i)}{w_i\sigma_i}\Big)} \times \\
     \prod_{i=1}^m \Phi\Big( \frac{g - w_i(\mu_i - z_i)}{w_i\sigma_i}\Big),
    \label{eq:g_pdf2}
\end{dmath}
where $\phi(\cdot)$ and $\Phi(\cdot)$ represent the probability density and cumulative density functions of the standard normal distribution, respectively. The distribution in Equation (\ref{eq:g_pdf2}) has a closed-form expression but it is not Gaussian distributed. Therefore, a closed-form expression of acquisition functions, which rely on the Gaussian assumption of the function cannot be used. A possible solution to this problem is to approximate the distribution with Generalised extreme value theory \cite{GEV,Haan2006}. The distribution can be described with Type 1 distribution in Generalised value distributions. Specifically, we use a Gumbel distribution \cite{GumbelBO} for approximating the scalarising function:
\begin{align*}
    g \sim  \text{Gumbel} \ (\alpha,\beta) 
\end{align*}
where $\alpha$ and $\beta$ are location and scale parameters, respectively. The probability density function is:
\begin{align*}
    p(g_i|\alpha,\beta) =& \frac{1}{\beta}e^{-(t_i + e^{-t_i})} \enspace \text{for} \enspace i =1,\ldots,N 
\end{align*}
where $t_i =\frac{g_i - \alpha}{\beta}$. We can estimate the parameters by maximising the following log-likelihood function
\begin{align*}
    LL(\alpha,\beta) = N \log \frac{1}{\beta} - \sum_{i=1}^N t_i - \sum_{i=1}^N e^{-t_i},
\end{align*}
where $N$ is the number of samples drawn from Equation (\ref{eq:g_pdf1}) or (\ref{eq:g_pdf2}).
The estimated parameters are:
\begin{align*}
    \alpha,\beta =&\argmax_{\alpha,\beta}\prod_{i=1}^N p(g_i|\alpha,\beta) \enspace \text{Or} \\ 
    \alpha,\beta =& \argmax_{\alpha,\beta} \Big( N \log \frac{1}{\beta} - \sum_{i=1}^N t_i - \sum_{i=1}^N e^{-t_i} \Big)
\end{align*}
After taking the partial derivatives, the parameters can be estimated by solving the following two equations:
\begin{align*}
    \beta =& \bar{g} - \frac{\sum_{i=1}^N g_i e^{-\frac{g_i}{\beta}}}{\sum_{i=1}^N e^{-\frac{g_i}{\beta}}} \\
    \alpha =& -\beta \log \Big( \frac{1}{N} \sum_{i=1}^N  e^{\frac{-g_i}{\beta}}\Big)
\end{align*}
Once the parameters are known, we can use the approximated distribution in estimating the EI: 
\begin{align*}
    \alpha_{EI} = \int_{-\infty}^{g'(\bx)} \max (0, g'(x) - g) \ dg
\end{align*}
One major drawback is that the EI does not have a closed-form expression. Therefore, we use Monte Carlo for estimating the EI. To summarise, we first approximate the distribution of TCH using Gumbel distribution and then draw samples from the distribution to estimate EI. For the two-objective example, the resulting TCH predictions and uncertainty estimates after building independent Gaussian process models on objective functions are shown in Figure \ref{fig:demo_4}. We also show the landscape of the expected improvement in the figure and can see that the optimal location is different from the mono-surrogate approach. The resulting decision variable is then evaluated with underlying objective functions and as can be seen, the objective function values lie on the Pareto front. This demonstration on an easy one-dimensional two objective optimisation problem shows that the multi-surrogate approach is better and finding an appropriate distribution of the scalarising function is important. 



\section{Numerical Experiments}
We investigate the performance of the multi-surrogate approach on standard DTLZ \cite{Deb2005b} benchmark and a real-world Free-Radical Polymerisation problem \cite{Chugh2014}. We compare it to the ParEGO as the mono-surrogate approach and multi-surrogate expected hypervolume improvement (EHVI) \cite{Emmerich2006,Yang2019}. We implemented all different approaches and used the same settings wherever possible for a fair comparison. The numerical settings used are as follows:
\begin{itemize}
    \item Problems: DTLZ (2, 5 and 7), Free-Radical Polymerisation
    \item Number of objectives: 2 and 3
    \item Number of decision variables: 4 and 5
    \item Number of independent runs = 11
    \item Size of the initial data set: 10 $\times$ number of decision variables
    \item Maximum number of function evaluations: 30 $\times$ number of decision variables 
    \item Kernel: Squared exponential (or Radial basis function, Gaussian) with automatic relevant determination
    \item Optimiser to maximise acquisition functions: Genetic Algorithm
    \item Optimiser to maximise marginal likelihood in Gaussian process: BFGS with 10 restarts 
    \item Performance indicator: Hypervolume 
\end{itemize}

\begin{figure*}[t]
    \centering
    \includegraphics[scale=0.25]{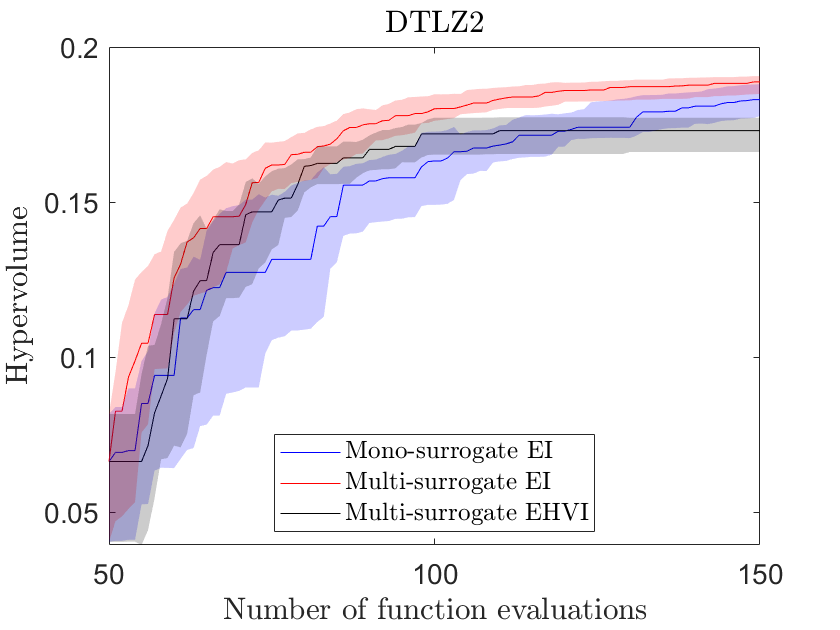}
    \includegraphics[scale=0.25]{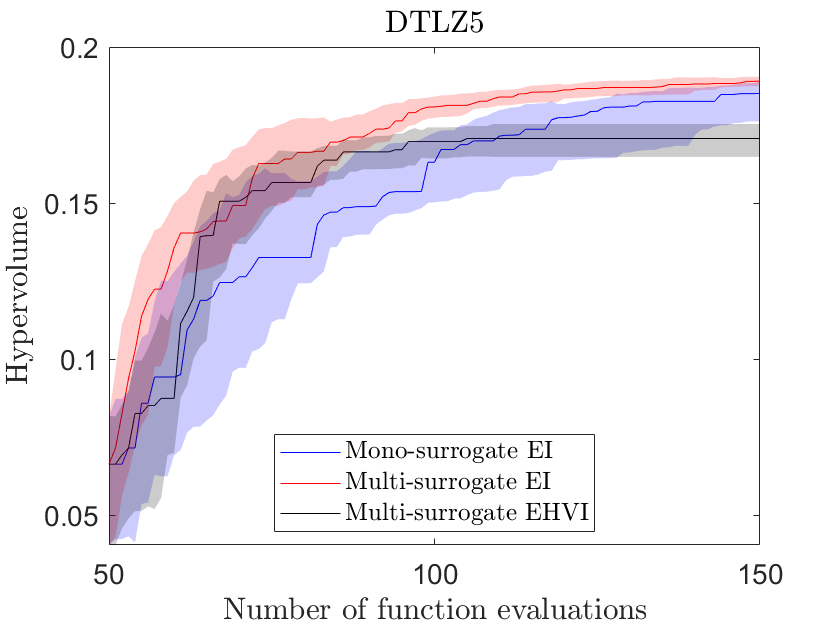}
    \includegraphics[scale=0.25]{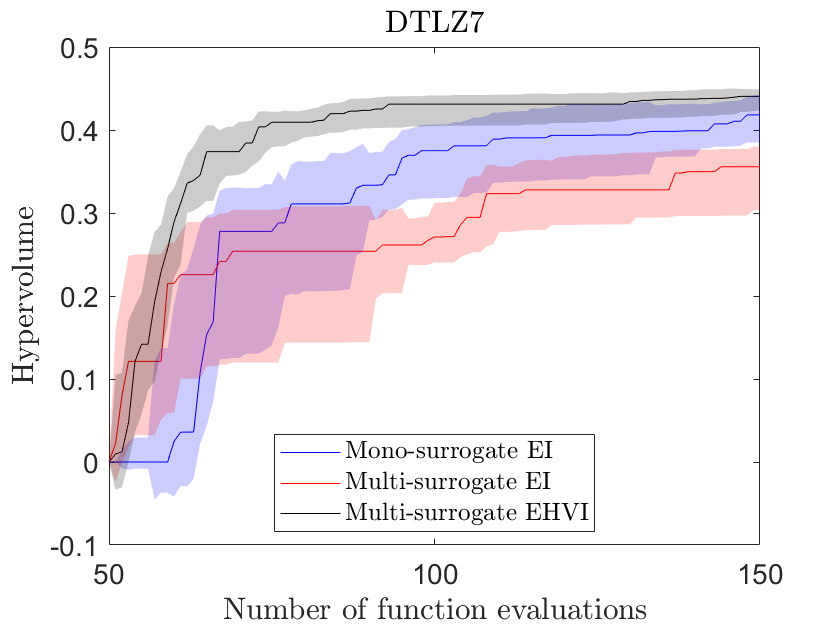}
    \caption{Hypervolume with the number of function evaluations on DTLZ problems with two objectives and five decision variables. The solid line is the median and shaded region is 95\% confidence interval.}
    \label{fig:hv_dtlz}
\end{figure*}

\begin{figure*}[t]
    \centering
    \includegraphics[scale=0.25]{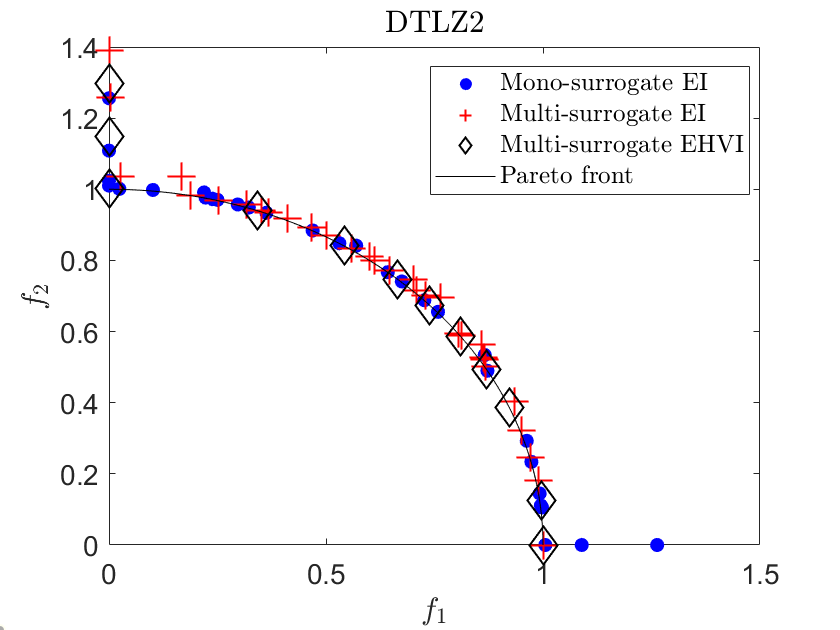}
    \includegraphics[scale=0.25]{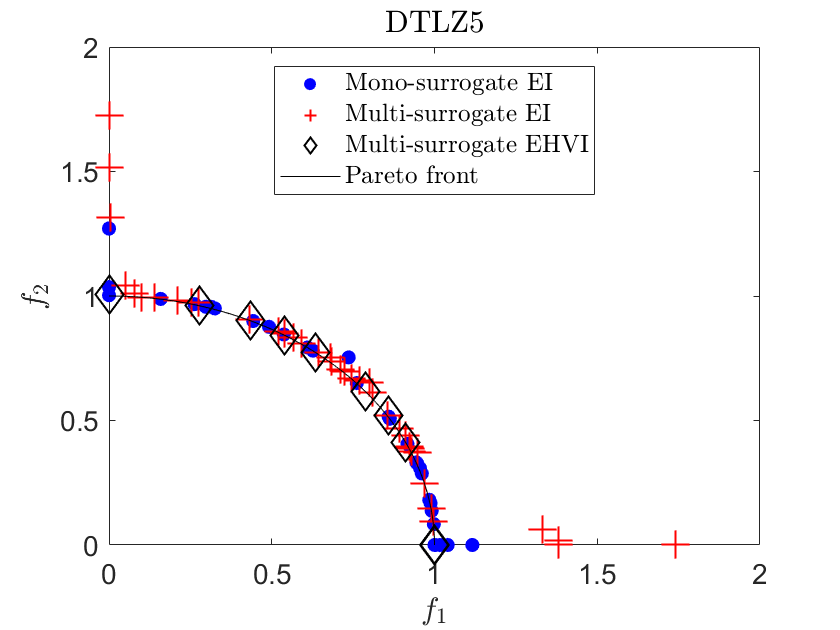}
    \includegraphics[scale=0.14]{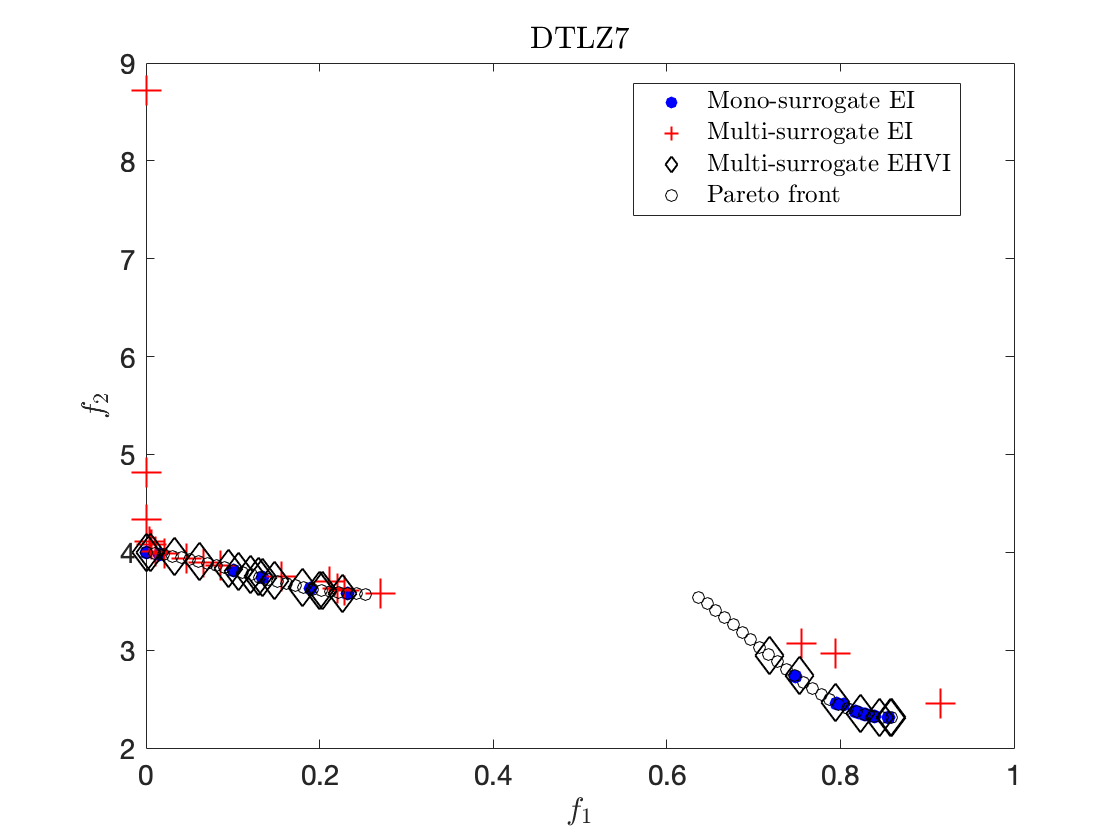}
    \caption{Approximated Pareto fronts of the run with the median hypervolume value.}
    \label{fig:pf_dtlz}
\end{figure*}

The hypervolume with the number of function evaluations for DTLZ problems with five decision variables and two objectives are shown in Figure \ref{fig:hv_dtlz}. As can be seen, the multi-surrogate approach outperformed the other two approaches. The performance of the mono-surrogate approach can be explained with our analysis in the previous section. In the multi-surrogate EHVI approach, the hypervolume contributions of solutions dominated by the current approximated Pareto front is zero, which can make it difficult for the algorithm to converge. The same concern in using EHVI was also mentioned in \cite{Alma2017}. The multi-surrogate approach did not perform well on the DTLZ7 problem. The problem has a disconnected Pareto front and separable objective functions. The performance of the multi-surrogate approach can be related to the shape of the Pareto front and is worth exploring as the future research direction. The approximated Pareto fronts with three different approaches of the run with the median hypervolume are shown in Figure \ref{fig:pf_dtlz}.



\begin{figure}[t]
    \centering
    \includegraphics[scale=0.16]{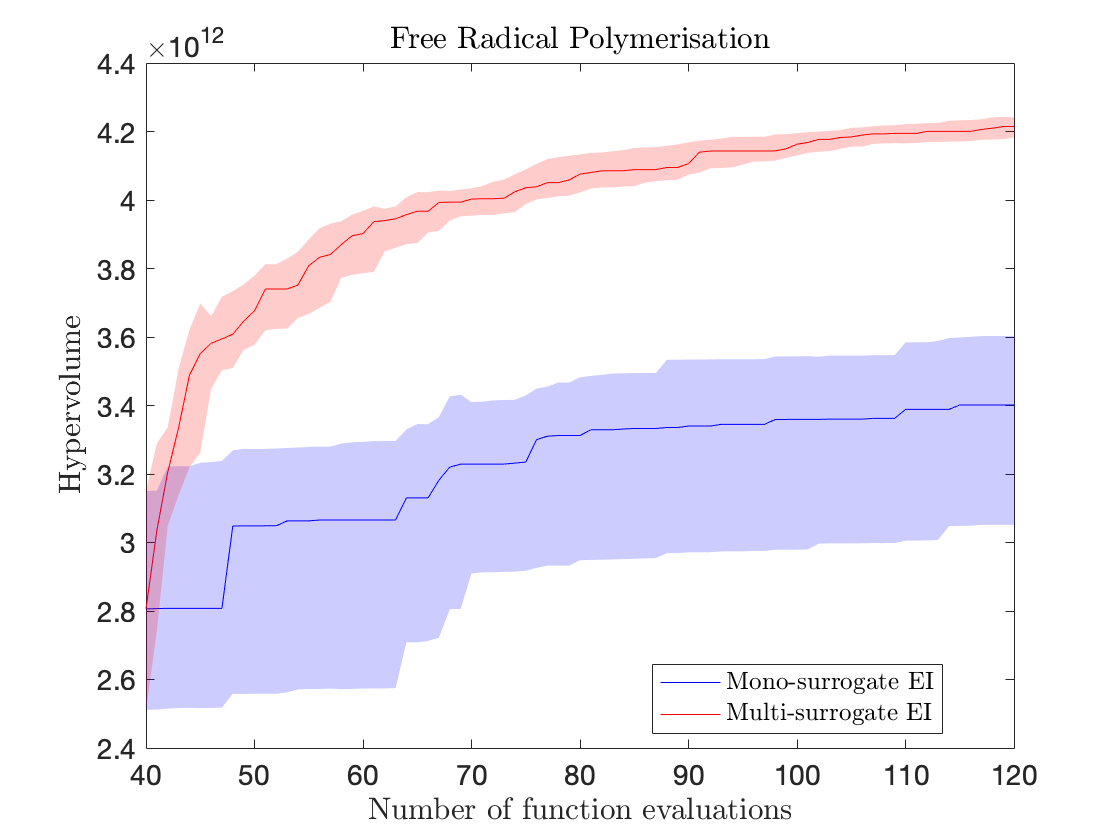} 
    \includegraphics[scale=0.16]{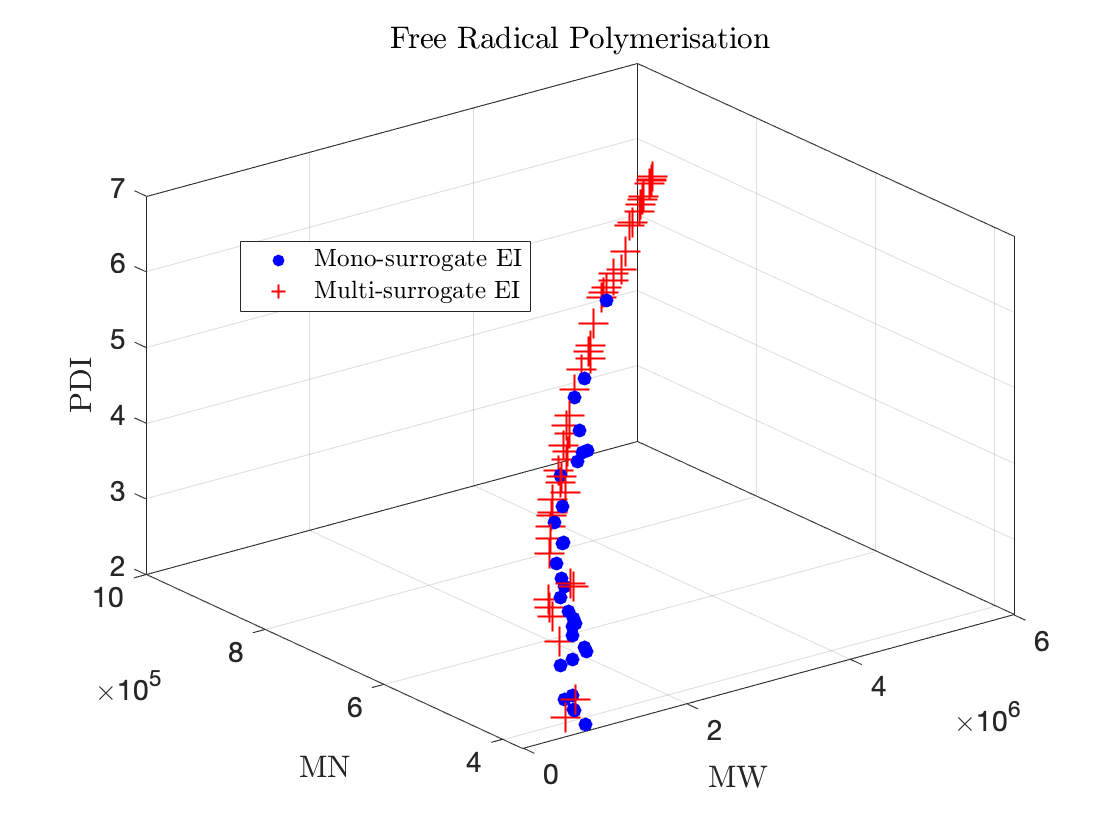}
    \caption{Hypervolume with the number of function evaluations on the Free-Radical Polymerisation problem (top figure). The approximated Pareto fronts, MW and MN are to be maximised and PDI is to be minimised.}
    \label{fig:hv_pf_FRP}
\end{figure}

\begin{figure}[t]
    \centering
    \includegraphics[scale=0.2]{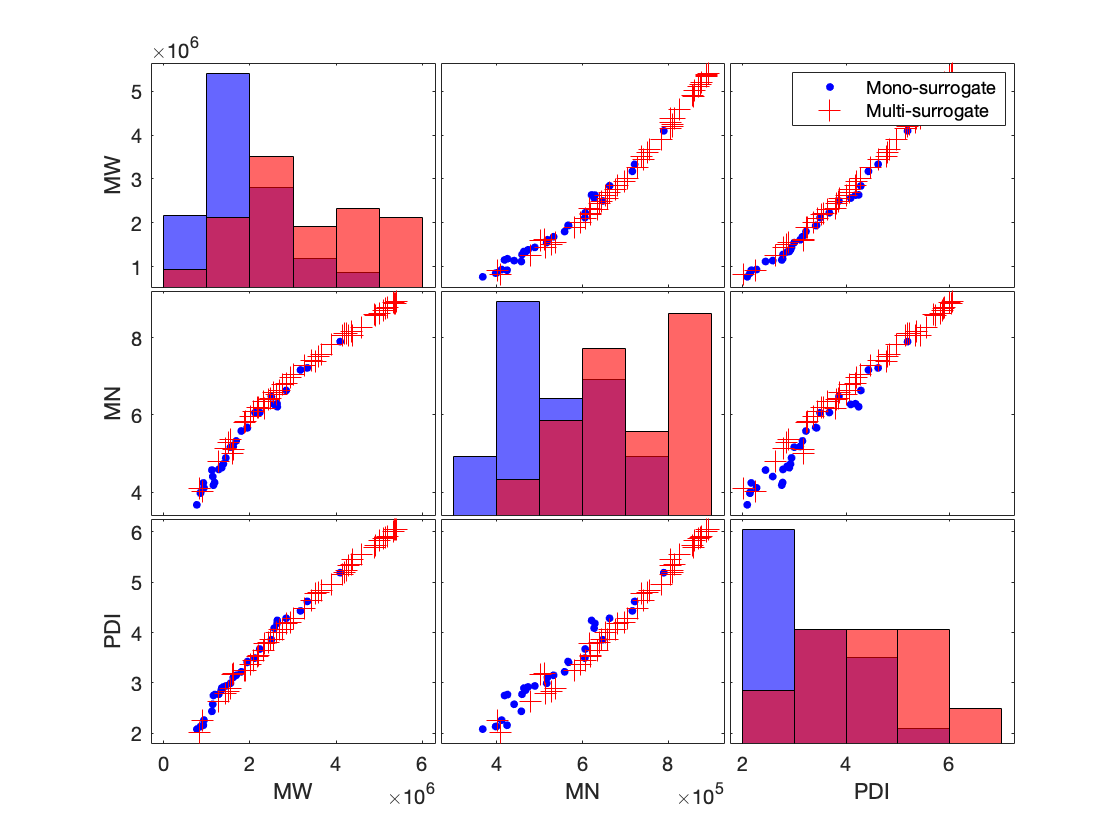}
    \caption{Two dimensional scatter plots between different objectives in the Free-Radical Polymerisation problem.}
    \label{fig:pf_FRP}
\end{figure}

The real-world problem is Free-Radical Polymerisation problem. The problem is the manufacturing of Polyvinyl Acetate polymer. The polymer is manufactured in a batch reactor and the process can be modelled with a series of ordinary differential equations. These equations are stiff and therefore solving these equations can be computationally expensive. The computation time varies from 30 seconds to 20 minutes for one evaluation. The problem involves four decision variables: monomer concentration, initiator concentration, the temperature of the rector and polymerisation time and three objectives: maximise weight average molecular weight (MW), number average molecular weight (MN) and minimise polydispersity index (PDI). The hypervolume and approximated Pareto fronts after running mono-surrogate and multi-surrogate approaches are shown in Figure \ref{fig:hv_pf_FRP}. As can be seen, the multi-surrogate approach outperformed the mono-surrogate approach. We also show the two dimensional scatter plots between different objectives in Figure \ref{fig:pf_FRP}.

\begin{figure}
    \centering
    \includegraphics[scale=0.38]{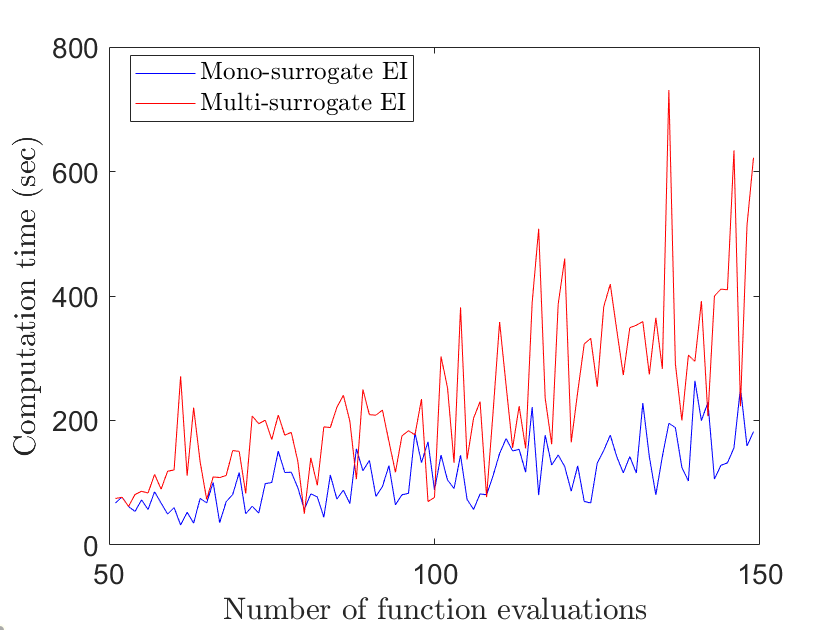}
    \caption{Computation time of mono- and multi-surrogate approaches on the DTLZ2 problem with two objectives and five decision variables}
    \label{fig:computation_time}
\end{figure}

The multi-surrogate approach relies on approximations and Monte Carlo simulations when estimating EI, which makes the optimisation process slower. The computation time on the DTLZ2 problem of one run with two objectives and five variables of both approaches is shown in Figure \ref{fig:computation_time}. As can be seen, the multi-surrogate approach is slower than the mono-surrogate approach. However, this computation time may not be significant in many real-world applications and may be negligible compared to the objective function evaluation. 

\color{red}

\color{black}

\section{Conclusions}
In this work, we presented a multi-surrogate approach in multi-objective Bayesian optimisation. In comparison to the mono-surrogate approach, the proposed approach did not use the surrogate model built on the scalarised function and did not assume the scalarised function distribution is Gaussian. We built independent Gaussian process models on the objective functions and showed that the weighted Tchebycheff as the scalarising function after building independent models for each objective function was not Gaussian. We approximated the scalarising function distribution with Type 1 of the Generalised value extreme value distributions and used Monte Carlo to estimate the acquisition function. For the demonstration of the potential of the multi-surrogate approach, we tested the approach on benchmark and real-world problems. The proofs and analysis of the results indicated that one should use an appropriate distribution of the resulting scalarising function in multi-objective Bayesian optimisation.

Future work will include testing on other scalarising functions e.g.\ hypervolume improvement, dominance rank, penalty boundary intersection and scalarising functions used in the evolutionary multi-objective optimisation algorithms. As mentioned, the multi-surrogate approach is more computationally expensive than the mono-surrogate approach. An alternative to alleviate the computation cost is to use the Laplace approximation for the distribution of the scalarising function. We provided initial calculations of the Laplace approximation in the Appendix. We plan to work on using these calculations in the future. Testing on a wide range of benchmark and real-world problems will also be in our future works. 

\begin{acks}
The author would like to thank Dr De Ath George at the University of Exeter, UK for his feedback on the paper.
\end{acks}

\section*{Appendix}
\subsection*{Laplace approximation calculations for bi-objective optimisation}

Let us denote $mi = w_i(\mu_i - z_i)$ and $si = w_i \sigma_i$, where $\mu_i$, $\sigma_i$, $z_i$ and $w_i$ are the $i^{th}$ element of the posterior mean vector, the standard deviation vector, the ideal objective vector and the weight vector, respectively. The distribution of $g$ mentioned in Equation (\ref{eq:g_pdf2}) can also be written as:
{\fontsize{6}{10}\selectfont
\begin{equation}
 p(g) = \frac{1}{4}\text{Erfc}\left[-\frac{g-\text{m1}}{\sqrt{2}\text{s1}}\right]\left(\frac{e^{-\frac{(g-\text{m1})^2}{2\text{s1}^2}}\sqrt{\frac{2}{\pi}}}{\text{s1}\text{Erfc}\left[-\frac{g-\text{m1}}{\sqrt{2}\text{s1}}\right]}+\frac{e^{-\frac{(g-\text{m2})^2}{2\text{s2}^2}}\sqrt{\frac{2}{\pi}}}{\text{s2}\text{Erfc}\left[-\frac{g-\text{m2}}{\sqrt{2}\text{s2}}\right]}\right)\text{Erfc}\left[-\frac{g-\text{m2}}{\sqrt{2} \text{s2}}\right] 
\end{equation}
}

The mode, $g_0$ can be calculated by taking the derivative of the $\log g$ i.e.\ $\frac{d \log g}{d g}$ =0:
\tiny{
\begin{dmath}
\frac{4 \text{s1}^2 \text{s2}^2+\sqrt{2 \pi } \left(-e^{\frac{(g-\text{m1})^2}{2 \text{s1}^2}} (g-\text{m2}) \text{s1}^3 \text{Erfc}\left[\frac{-g+\text{m1}}{\sqrt{2}
\text{s1}}\right]-e^{\frac{(g-\text{m2})^2}{2 \text{s2}^2}} (g-\text{m1}) \text{s2}^3 \text{Erfc}\left[\frac{-g+\text{m2}}{\sqrt{2} \text{s2}}\right]\right)}{\sqrt{2
\pi } \text{s1}^2 \text{s2}^2 \left(e^{\frac{(g-\text{m1})^2}{2 \text{s1}^2}} \text{s1} \text{Erfc}\left[\frac{-g+\text{m1}}{\sqrt{2} \text{s1}}\right]+e^{\frac{(g-\text{m2})^2}{2
\text{s2}^2}} \text{s2} \text{Erfc}\left[\frac{-g+\text{m2}}{\sqrt{2} \text{s2}}\right]\right)} =0
\end{dmath}}

\normalsize{The equation does not have a closed form expression for estimating the mode. Therefore, the mode can be estimated by maximising a posteriori with an iterative method e.g.\ Newton's method.}

\normalsize{Once the mode is known, we can estimate the second derivative at the mode:}
\tiny{
\begin{dmath}
    A = -\diff*[2]{\log g}{g}{g = g_0} = \left(-e^{\frac{(g-\text{m1})^2}{\text{s1}^2}} \pi  \text{s1}^5 \text{s2} \text{Erfc}\left[\frac{-g+\text{m1}}{\sqrt{2} \text{s1}}\right]^2+e^{\frac{(g-\text{m1})^2}{2
\text{s1}^2}} \text{Erfc}\left[\frac{-g+\text{m1}}{\sqrt{2} \text{s1}}\right] \left(\sqrt{2 \pi } \text{s1}^2 \text{s2} \left((g-\text{m2}) \text{s1}^2+3
(-g+\text{m1}) \text{s2}^2\right)+e^{\frac{(g-\text{m2})^2}{2 \text{s2}^2}} \pi  \left((g-\text{m2})^2 \text{s1}^4-\text{s1}^2 \left(2 (g-\text{m1})
(g-\text{m2})+\text{s1}^2\right) \text{s2}^2+(g-\text{m1}-\text{s1}) (g-\text{m1}+\text{s1}) \text{s2}^4\right) \text{Erfc}\left[\frac{-g+\text{m2}}{\sqrt{2}
\text{s2}}\right]\right)-\text{s1} \text{s2}^2 \left(8 \text{s1}^2 \text{s2}+e^{\frac{(g-\text{m2})^2}{2 \text{s2}^2}} \sqrt{2 \pi } \left(3 (g-\text{m2})
\text{s1}^2+(-g+\text{m1}) \text{s2}^2\right) \text{Erfc}\left[\frac{-g+\text{m2}}{\sqrt{2} \text{s2}}\right]+e^{\frac{(g-\text{m2})^2}{\text{s2}^2}}
\pi  \text{s2}^3 \text{Erfc}\left[\frac{-g+\text{m2}}{\sqrt{2} \text{s2}}\right]^2\right)\right)/\left(\pi  \text{s1}^3 \text{s2}^3 \left(e^{\frac{(g-\text{m1})^2}{2
\text{s1}^2}} \text{s1} \text{Erfc}\left[\frac{-g+\text{m1}}{\sqrt{2} \text{s1}}\right]+e^{\frac{(g-\text{m2})^2}{2 \text{s2}^2}} \text{s2} \text{Erfc}\left[\frac{-g+\text{m2}}{\sqrt{2}
\text{s2}}\right]\right)^2\right)
\end{dmath}
}

\normalsize{After estimating the mode and the second derivative at the mode, the distribution of $g$ can be written as:
\begin{align*}
    g \sim \mathcal{N}(g_0, A^{-1}),
\end{align*}}
where $g_0$ is the mean and $A^{-1}$ is the variance. These calculations allow us to use the closed form expression of acquisition functions e.g.\ expected improvement and probability of improvement. 

\subsection*{Results on DBMOPP}
\begin{figure}[t]
    \centering
    \includegraphics[scale=0.3]{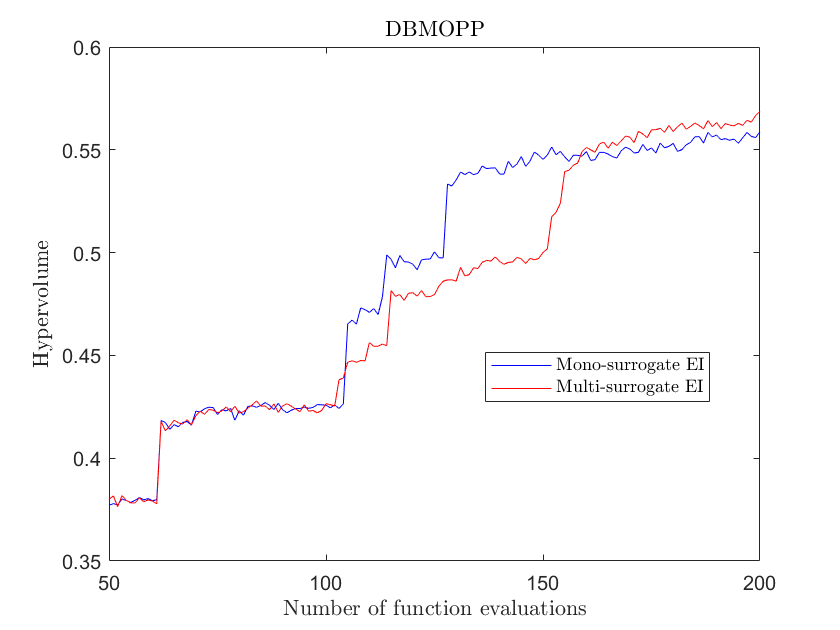}
    \caption{Hypervolume with the number of function evaluations on DBMOPP problems with five objectives and 10 decision variables.}
    \label{fig:hv_dbmopp}
\end{figure}

\begin{figure}[t]
    \centering
    \includegraphics[scale=0.3]{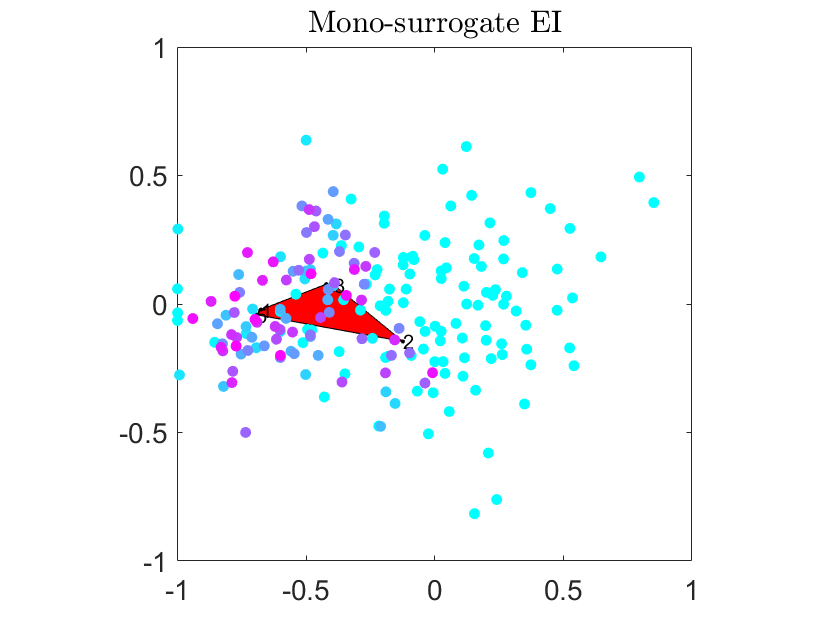}
    \includegraphics[scale=0.3]{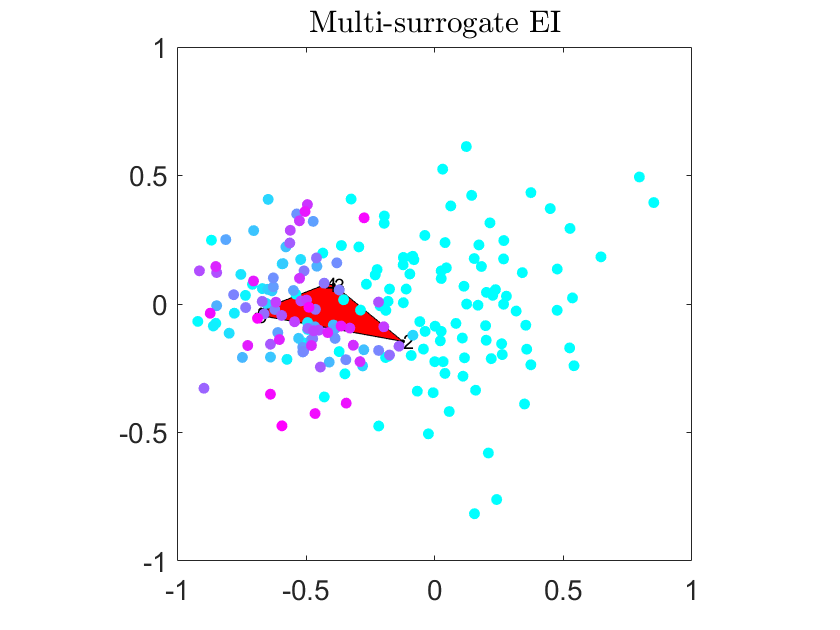} \\
    \includegraphics[scale=0.6]{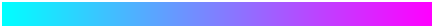}
    \caption{Trace iteration plots for a DBMOPP problem with five objectives and 10 decision variables. The colormap represents the iteration counter between 0 - 200.}
    \label{fig:pf_dbmopp}
\end{figure}

In addition to DTLZ, we tested the proposed approach on a single instance of DBMOPP \cite{Jonathan_DBMOPP,Fieldsend2021}. The results of hypervolume with five objectives and 10 decision variables are shown in Figure \ref{fig:hv_dbmopp}. We did not see a significant difference in the results of the two approaches. As DBMOPP problems are visualisable, the solutions can be projected to two dimensions. We plotted the solutions with the number of function evaluations in Figure \ref{fig:pf_dbmopp}. The solutions on or inside the convex hull (shown in red) represent the Pareto optimal solutions. As can be seen, both approaches converged closer to the Pareto front. This problem is relatively easier and does not have any complexity like local Pareto fronts, varying density and dominance resistance regions. It will be worth testing the multi-surrogate approach on complex DBMOPP problems.

\color{black}

\bibliographystyle{ACM-Reference-Format}
\bibliography{main}

\newpage

\end{document}